\documentclass{article}
\usepackage[super]{nth}
\usepackage{amssymb}
\usepackage{graphicx}
\usepackage{scrextend}
\title{Deepfake Detection using ImageNet models and Temporal Images of 468 Facial Landmarks}
\author{Christeen T Jose\\
Vellore Institute of Technology, Vellore, India\\
christeenjosethoomkuzhy@gmail.com}

\renewcommand{\today}{\ifcase \month \or January\or February\or March\or %
April\or May \or June\or July\or August\or September\or October\or November\or %
December\fi, \number \year}

\begin{document}
\maketitle
\begin{abstract}
This paper presents our results and findings on the use of temporal images for deepfake detection. We modelled temporal relations that exist in the movement of 468 facial landmarks across frames of a given video as spatial relations by constructing an image (referred to as temporal image) using the pixel values at these facial landmarks. CNNs are capable of recognizing spatial relationships that exist between the pixels of a given image. 10 different ImageNet models were considered for the study.
\end{abstract}
\providecommand{\keywords}[1]{\textbf{\textit{Index terms---}} #1}
\keywords{deepfake detection, deep learning, temporal images}
\section{Introduction}
Deep-fakes are fake videos that contain content generated using deep neural network architectures. The main machine learning methods used to create deepfakes are based on deep learning and involve training generative neural network architectures, such as autoencoders or generative adversarial networks (GANs). This capability of deep neural network architectures to produce fake videos has been widely exploited and continues to be misused for purposes such as propagation of misinformation, evidence tampering, political manipulation, fraud, non-consensual pornography, revenge porn, and acts against national security. Thus the development of Deep-fake Detection mechanisms has become an utmost need of the 21st century.

Our objective is to build a novel approach to solving the problem of deepfake detection, that would enable researchers to work with only a fraction of the available data  thus saving computational power and time, while providing accurate results. Deepfake detection mechanisms rely on identifying spatial and temporal inconsistencies in a given video, which requires the processing of a sequence of frames before any conclusions can be made. 

In our approach we would not be considering entire frames, but only the pixel values at each of the 468 facial landmarks on the face, thus considerably reducing the amount of input data. The novelty of our approach is that we will be modelling temporal relations that exist in the movement of 468 facial landmarks as spatial relations by constructing an image from facial landmark data across frames. The conventional approach to learning temporal relations is to use a Recurrent Neural Network (RNN), but by mapping temporal data as spatial data we can avoid using an RNN, thus saving computational power and time taken for training. 

Each pixel in the newly constructed temporal image corresponds to the RGB values at the pixel location of each of the 468 facial landmarks for each of the frames in the considered sequence. Convolutional Neural Networks are capable of recognizing spatial relationships that exist between the different pixels of a given image and this has been the major motivation behind our idea to build temporal images. Temporal images are image files that provide a visual representation of temporal data. Temporal data can be defined as the collection of values across time for a property of an object under observation. Such images can map temporal relations that exist within the temporal data as spatial relations.
\section{Literature review}
Democracy, national security, and society are under significant threat with the emergence of deepfake videos. Deepfake videos targeted at world leaders, army chiefs, or corporate titans can result in a constitutional crisis, civil unrest, or global stock manipulation, through the spread of misinformation. \cite{agarwal2019protecting} Deepfakes can damage international relations, manipulate the economy, elections, democratisation of policy-making, harm military or intelligence operations, exploit social divisions, and cause widespread mistrust. There is also the threat of the creation of deepfakes with cruel and malicious intentions as weapons meant for terrorizing and inflicting pain upon targets. \cite{chesney2019deep} Detection of such fake face videos has become a pressing need for the research community of digital media forensics. \cite{li2018ictu}

A study from 2015 had showed that humans could easily be fooled by manipulated digital images and this was before the emergence of deepfakes. The researchers had conducted a study to analyse the ability of humans at identifying forged images. 383 individuals had taken part in this study and there were a total of 17,208 responses. In their study, it was observed that modified images were correctly identified only 46.5\% of the time. The overall accuracy at differentiating between modified and original images was 58\%. \cite{schetinger2017humans}

Presented in 2016, Face2Face could create videos of a target person given his/her video and a source video of another individual for facial reenactment by the target. \cite{thies2016face2face} Further, in 2017, it was shown that fake videos of a target individual could be created with just a single image of the target along with the source video of another individual. The researchers had demonstrated this on various still portraits from the internet. \cite{averbuch2017bringing} 2017 also saw the application of recurrent neural networks to map raw audio features with mouth shapes for the creation of photorealistic lip-sync videos. \cite{suwajanakorn2017synthesizing}

A 2002 study had previously shown that individuals could be recognized using their facial behavior alone and the stability of these individual differences over time. Facial action units can be used for the interpretation of psychological states and recognizing individuals as they convey unique information about an individual's identity. \cite{cohn2002individual} Facial action units are used for the description of facial movements and have been derived from an anatomical basis. \cite{ekman1976measuring} Further, a 2010 study, was able to demonstrate how individuals could be identified based on their upper body motion features. \cite{williams2010body}
Although, methods for establishing the authenticity of digital images and videos have been greatly studied in the past even before the emergence of deepfakes \cite{farid2016photo}, deepfake videos pose challenges to traditional media forensics techniques which rely upon artifacts left from sensor noise, CFA interpolation, double JPEG compression, lighting conditions, shadows, reflections and the consistency of meta-data. \cite{li2018ictu}

Deepfakes have been grouped into three categories namely face-swap, lip-sync and puppet-master based on the regions that have been artificially constructed. Face-swap deepfakes are constructed by replacing only the facial region of the head of the person appearing, in the video whereas, in lip-sync deepfakes, only the mouth region is modified to make it consistent with an arbitrary audio recording. From the definition of lip-sync deepfakes, we can infer the role of fraudulent/fake audio recordings in making believable deepfake videos. In the case of puppet-master deepfakes, the entire head of a target person is created artificially, including head movements, eye movements, and facial expressions from a performer acting out what the puppet (target person) should say and do. \cite{agarwal2019protecting}

The first forensic method specifically targeted at the detection of AI-generated fake face videos was proposed in 2018. The method relied upon the lack of physiological signals which are intrinsic to human beings, in these videos. The lack of eye blinking was the focus of this early approach, although the authors mention the possibility of considering other spontaneous and involuntary physiological activities such as breathing, pulse, and eye movement. The alignment of facial regions to a unified coordinate space helped to avoid distractions in facial analysis introduced by the movement of the head and changes in the orientation of the face between video frames. For capturing the phenomenological as well as temporal regularities that exist in the process of eye blinking, the authors used a Long-term Recurrent Convolutional Neural Networks (LRCN) trained as a binary classifier for distinguishing between open and closed states of the eye, as eye blinking is a temporal process. LRCN is a combination of a convolutional neural network (CNN) and a recursive neural network (RNN). The model showed promising performance in detecting deepfake videos. \cite{li2018ictu}

Shortly after this technique was made public, deepfake synthesis systems started incorporating components for blinking thus rendering the forensic technique less effective against newer deepfakes. The same team further developed a more advanced technique based on distances between the features around the facial region and features in the central part of the facial region from the estimated 3D head pose. \cite{agarwal2019protecting} This was based on their observation that deepfakes contain errors introduced during splicing of a synthesized face region onto an original image as the target face and original face will have mismatched facial landmarks and that these errors can be revealed in a 3D head pose estimation of the facial region from the image. Individual frames from videos were considered. An important point to note here is that the study did not consider all of the 68 facial landmarks that were extracted and decided to focus only on those around the face and in the central region. The differences between the central and outer facial regions in the estimated head pose were used to build a feature vector for training an SVM classifier to distinguish between real and fake images. \cite{yang2019exposing} Although this method was effective against face-swaps, it was not found to be effective at detecting puppet-master or lip-sync deepfakes. \cite{agarwal2019protecting} The same team had published another method for exposing deepfakes based on face warping artifacts. \cite{li2018exposing}

In 2018 two neural networks namely Meso-4 and MesoInception-4 each with a low number of layers, were introduced for the detection of realistic forged videos. Meso-4 had 4 convolutional layers whereas MesoInception-4 replaced the first two convolutional layers with a variant from the inception architecture. The study focused on the mesoscopic properties of images. Microscopic analysis based on image noise won't apply for compressed videos, meanwhile, the human eye struggles to distinguish between real and forged images at a higher semantic level. An important observation made by the researchers while visualizing the layers and filters of these networks was the paramount role played by the eyes and mouth as these regions were detailed by positive-weighted neutrons, whereas the background (region excluding face) was detailed by negative-weighted neutrons. The study observed detection rates being as high as 98\% and 95\% for Deepfake videos and Face2Face videos respectively. \cite{afchar2018mesonet}

2018 also saw the application of a capsule network for the detection of forged images and videos as well as replay attacks. The inputs to the capsule network were latent features extracted using a part of a pre-trained VGG-19 network. \cite{nguyen2019capsule}

Another study from 2018 aimed to detect deepfakes using a convolutional LSTM structure by processing frame sequences looking for intra-frame and temporal consistencies. The convolutional LSTM model has two components namely a CNN part for feature extraction and an LSTM part for temporal sequence analysis. The InceptionV3 model was adopted for the CNN part. Sequences of length 20, 40, and 80 frames were considered and all showed promising test accuracies of above 96\% at successfully detecting deepfake videos. The study considered sequences of continuous frames and could detect deepfakes with less than 2 seconds of video footage. \cite{guera2018deepfake}

A 2018 study focused on the classification of images as real or GAN-generated based on GAN fingerprints. In the case of GAN-generated images, the study further focused on identifying their sources i.e. the GAN models used for synthesis. The study showed the GAN models have distinct fingerprints and that even minor differences in training can result in different fingerprints. Thus a fine-grained model can be used for authentication. Finetuning of fingerprints was found to be an effective immunization against deterioration of fingerprints by various image perturbation attacks. \cite{yu2019attributing}

Another forensic technique involving violations of physical signals was proposed in 2019, it was particularly aimed at protecting world leaders and other celebrities from the potential misuse of the large volumes of facial data that are publicly available. The study was able to show that individuals exhibited relatively distinct facial and head movement patterns while speaking, which were violated by all 3 categories of deepfakes. In the case of puppet-master and lip-sync deepfakes, the head movements and mouth region movements come from the impersonator respectively. The study used this new soft bio-metric to distinguish between real and fake videos of high-profile individuals. Such soft biometric models have been used in the past for identifying individuals even before the emergence of deepfakes. The major highlight of the study from 2019 and proposed soft-biometric was that it was able to successfully detect all 3 categories of deepfakes. The study extracted 10-second clips and saved them at 30 fps using a sliding window approach across videos of persons of interest (POIs) subject to the conditions that the only face visible in the clip belongs to the POI, the POI is speaking during the entire clip and the camera remains relatively stationary. The study then tracked facial and head movements, extracted specific action units, and then used a one-class support vector machine (SVM) for distinguishing target individuals from other individuals, deep-fake impersonators, and comedic impersonators. \cite{agarwal2019protecting}

A new forensic technique for the detection of face-swap deepfakes was presented in a 2020 paper. The paper also claimed that the most common deepfake detection approaches relied upon the presence of low-level pixel artifacts left-over from the synthesis process, that these approaches were vulnerable to simple countermeasures such as transcoding and resizing and that these approaches fail in generalizing to new techniques for that may arise deepfake synthesis. \cite{agarwal2020detecting} The same authors had mentioned other countermeasures for destroying artifacts such as additive noise and recompression. \cite{agarwal2019protecting} The technique combines the temporal behavioral biometric of facial behaviors (facial expressions and head movements) with the static biometric of facial identity as the former belongs to the original/target individual whereas the latter belongs to another individual. These inconsistencies can be revealed upon a comparison against a set of authentic reference videos, although the authors have considered the possibility that this technique might not work against lip-syncing deepfakes. Each of the reference and testing videos was saved at 25fps and an FFmpeg quality of 20. This constant frame rate allowed each of the videos to be partitioned into overlapping clips of 4-second duration each. (100 frames) with a sliding window of 5 frames. Spatiotemporal features across 4 seconds were captured by stacking static FAb-Net features. This technique does not require a specific model for each individual and allows for a generic model that can be trained on one group of individuals and could be generalized to previously unseen individuals as well. The VGG 16-layer Convolutional Neural Network architecture was used for face recognition. This combination of facial behaviors and facial identity metrics was shown to be highly effective at detecting face-swap deepfakes. \cite{agarwal2020detecting}

Meanwhile, in 2020, deepfake evasion mechanisms were also being greatly studied. It was shown that flipping the lowest bit value of each pixel in an image/frame could drastically reduce the performance of forensic classifiers. Such evasion mechanisms can lower the accuracy of forensic classifiers close to 0\%. Incorporating single noise patterns in the synthesizer was also found to significantly lower the forensic classifier performance. Standard laundering attacks include resizing, recompression, blurring, and adding white noise. These attacks aim to imperceptibly modify fake images so that a forensic classifier misclassifies them as real. There also arises the possibility that real images are modified imperceptibly for them to be mislabeled as being fake by a forensic classifier. The paper also points out the false sense of security that is provided by the deployment of vulnerable forensic classifiers as fake pictures would now have credibility provided by these forensic classifiers. \cite{carlini2020evading}

2020 also saw a new approach to deepfake detection based on disparities in color components between DNG and real images. Deepfake images were referred to as Deep Network Generated (DNG) images by this study and it was observed that DNG images were distinguishable from real images in terms of chrominance components. The generative models behind the synthesis of DNG images attempt to imitate real images that are in the RGB color space. It was found that these statistical properties for DNG images and real images were different under HSV and YCbCr color spaces. These disparities were found to be more distinct in the residual domain. The study developed a low-dimension feature set for capturing color image statistics. The feature set consisted of co-occurrence matrixes that had been extracted from the residual images for several color components. The method could accurately identify DNG images and outperformed other existing methods when there was a mismatch between training and testing data. The method also showed a good performance in one-class classification as well, when the GAN model behind the synthesis of the DNG image was unknown, where only real images were used for training. \cite{li2020identification}

Another study from 2020 showed that some common systemic flaws are shared between all CNN-generated images and that forensic classifiers could easily generalize between different models used for synthesis. Detectable fingerprints are retained in all CNN-generated images and can be used for distinguishing such images from real images. \cite{wang2020cnn}

In 2020 a new technique for the detection of deepfakes based on phoneme-viseme mismatches was introduced. This allowed for the detection of small spatial and temporal manipulations in videos as a new kind of deepfakes that contained only a small region of replaced synthesized content for the mouth region, was emerging. The study focused on visemes that required a completely closed mouth for the pronunciation of their corresponding phonemes, which was violated in many deepfakes. The method was able to successfully detect state-of-the-art lip-sync deepfakes. \cite{agarwal2020detectingPV}

DeepRhythm is a more advanced deepfake detection technique that monitors heartbeat rhythms for disruptions. Remote visual photoplethysmography (PPG) allows for determining the heartbeat rhythm by monitoring minute periodic changes in skin color. \cite{qi2020deeprhythm} Another study used PPG cells extracted from real and fake videos for training a classification network for determining the generative model behind each video. This was the first method to make use of biological signals for deepfake source detection. The method was tested on different window sizes of 64, 128, 256, and 512 frames for training a CNN model with 3 VGG blocks on the FaceForensics++ dataset, from which a window size of 64 frames was found to be optimal. The method could identify deepfake videos with an accuracy of 97.29\% and the source model with an accuracy of 93.39\% on the FaceForensics++ dataset. \cite{ciftci2020hearts}
\section{Research method}
Our solution is to construct temporal RGB images from a sequence of video frames where each pixel in the temporal image corresponds to the RGB values at the location of each of the 468 facial landmarks for each of the frames in the sequence, following the constraint that exactly one and the same face is present throughout the sequence without any discontinuities. A temporal image can be defined as the output of a data compression technique that generates an output image using the values at select pixels/data-members across multiple raw input images/objects that share a temporal relation. A temporal image is used to map temporal relations as spatial relations. These temporal images are then provided to a Convolutional Neural Network (CNN) for training along with labels that indicate whether each temporal image was constructed from a deep-fake video or not. The hypothesis behind this project is that the neural network will be able to recognize patterns from the temporal images for deepfake detection.

We believe that significant results can be achieved by training an ImageNet model on a temporal image dataset provided the existence of consistent temporal relations within the images of the temporal dataset. A temporal image dataset is a collection of temporal images along with their target values. Every temporal image in a temporal image dataset will have a target value associated with it.

The project was broken down into 8 major tasks. The first task was to carry out an extensive literature survey to understand the evolution of deepfake detection mechanisms. Further we started defining our concept of “temporal images” by coming up with algorithms and formulations in task 2, as inferences were made from the survey. In task 3, we requested access to select deepfake detection datasets namely UADFV \cite{li2018ictu}, Celeb-DF \cite{Celeb_DF_cvpr20}, FaceForensics \footnote{Used compressed version} \cite{roessler2018faceforensics}, DFD (Google) \footnote{\label{c23}Used c23 quality} \cite{DDD_GoogleJigSaw2019}, Celeb-DF-v2 \cite{Celeb_DF_cvpr20} and FaceForensics++ \footref{c23} \cite{roessler2019faceforensicspp}. Once we were granted access to these datasets we moved on to task 4, where we generated temporal images using each of these datasets. Task 5 was to apply pre-processing on the generated temporal images and was pipelined to be carried out as soon as a dataset had completed task 4, in order for saving time. Task 6 was to select the optimal parameters for ‘MobileNet’ and was pipelined with task 5 to be carried out as soon as a dataset had completed pre-processing. MobileNet having a smaller number of parameters was chosen for our initial experiments as it could yield quick results. After task 6 had concluded the next task was to conduct a comparative study on multiple ImageNet models using the optimal parameters obtained from task 6. The ImageNet models considered for the comparative study were: MobileNetV2 \cite{sandler2018mobilenetv2}, Xception \cite{chollet2017xception}, InceptionResNetV2 \cite{szegedy2017inception}, InceptionV3 \cite{szegedy2016rethinking}, DenseNet121 \cite{huang2017densely}, EfficientNetB0 \cite{tan2019efficientnet}, MobileNet \cite{howard2017mobilenets}, ResNet50 \cite{he2016deep}, VGG16 \cite{simonyan2014very} and VGG19 \cite{simonyan2014very}.

The dataset was split into train, valid and test sets before the application of any pre-processing steps so as to ensure that all the temporal images generated from a video belong to exactly one of train, valid and test sets. This was to avoid the model from overfitting on the dataset, if split at the frame/ temporal image level. This was in accordance with our original experiments where we observed very high test accuracies and steep learning curves due to overfitting. The split ratio chosen was 0.8 : 0.1 : 0.1, this implied that 80\% of the videos would be allotted for training and 10\% each for validation and testing.

Python was the programming language of our choice. MediaPipe was used to obtain the 468 facial landmarks, OpenCV was used for image processing, Tensorflow was used for training and testing on ImageNet models and Matplotlib was used for visualization.

During the study we were restricted to a RAM size of 12.68 GB. Therefore to avoid RAM Out Of Memory Error, resized temporal images were also considered.

MobileNet was the first ImageNet model we tried, and the model required input pixels to be scaled between -1 and 1. This introduced a new problem as the size of the temporal image dataset was increased by a factor of 4 after pre-processing, as 8 bit unsigned integer values were being replaced by float values at each pixel for each temporal image. We were also storing the target label values for each temporal image in each of the train, test, and validation sets. Thus standardisation and normalisation posed a challenge to using the datasets. In order to avoid Out Of Memory errors and to use the available datasets, it was decided to avoid any model specific pre-processing.

We decided to conduct our study on only the UADFV dataset due to spatial and temporal constraints. UADFV and Celeb-DF were the only two datasets that we could handle with our limited RAM. These were the smallest of the 6 considered datasets. Celeb-DF, the larger of the two, took a very long time to complete any operation and hence was avoided.

Noisy learning curves presented another challenge. The solution was to not restore weights from the best epoch and to use a higher value of 50 for patience to ensure that the model does not stop training around the global optimum for one metric.
\subsection{Temporal Images}
From a single face, we could obtain 468 facial landmarks. We could create an RGB image where each row is dedicated to the pixels at the 468 facial landmarks from each frame in a sequence of 468 such frames, subject to the constraint that exactly one and the same face is present throughout the sequence without any discontinuities. But to limit computational costs the dimensions for the temporal image were chosen to be 116x117. 117 was chosen as the width for the temporal image as it was a factor of 468, thus enabling all the pixels at the 468 facial landmarks from a single frame to occupy exactly 4 rows in the temporal image.  116 was chosen as the height for the temporal image as it is a multiple of 4 and close to 117. Thus a single temporal image could contain 468 pixels from 29 different frames ($29*4=116$). 

We make the assumption that the CNN will be able to figure out the sequential nature from the arrangement of rows of pixels, as the first four rows of the temporal image will be occupied by facial landmarks from the first frame in the sequence of 29 frames, and the second four rows will be occupied by facial landmarks from the second frame and so on. Therefore we can use a sliding window with window size set to 29 for the generation of temporal images. A single window now considers only 0.966 seconds of data from a 30fps video. In order to increase the duration considered by a window, we must skip some frames.

If we are to consider every \nth{15} frame, then a single window will require 14.033 seconds ($421/30$) of footage or  421 frames ($1 + (29-1)*15$).

But a lot of videos in publicly available datasets such as Celeb-DF have a duration much lower than that, and hence those videos will not be utilized. Therefore a much smaller value of 7 was chosen for I. Now a single window considers 6.566 seconds of footage or  197 frames.

For generating temporal images from a given video, we apply 7 sliding windows where each of the $i$ sliding windows ($0<=i<=6$)  has a window size of 29 frames and considers every \nth{7} frame shifted by $i$ frames in the video sequence.

We considered a total of 6 different deepfake detection datasets. The details of each dataset has been provided in table \ref{Datasets table}.

Videos that either didn't follow the constraint or were of shorter duration than 6.566 seconds  were not considered while generating temporal images. An example for such problematic videos is “id10\_id13\_0006.mp4” found in the “Celeb-synthesis” subdirectory of the Celeb-DF dataset, which contains multiple faces and is only 5 seconds long.

\begin{table}[ht]
\begin{center}
\caption{Datasets used for deepfake detection}
\label{Datasets table}
\begin{tabular}{|c|c|c|c|c|}
\hline
Dataset                    & Total     videos & Real     videos & Deepfake videos & Total number of frames \\\hline
UADFV                      & 98               & 49              & 49              & 34,320                 \\\hline
Celeb-DF                   & 1,203            & 408             & 795             & 4,88,278               \\\hline
FaceForensics              & 2,008            & 1,004           & 1,004           & 10,37,776              \\\hline
DFD (Google)               & 3,431            & 363             & 3,068           & 25,58,073              \\\hline
Celeb-DF-v2                & 6,529            & 890             & 5,639           & 24,75,558              \\\hline
FaceForensics++            & 6,000            & 1,000           & 5,000           & 28,48,789              \\\hline
\end{tabular}
\end{center}
\end{table}

Table \ref{Datasets 116x117} provides the details of the temporal images generated from each of the 6 different datasets.

\begin{table}[ht]
\begin{center}
\caption{Details of generated temporal images for dimensions (116, 117, 3)}
\label{Datasets 116x117}
\begin{tabular}{|c|c|c|c|}
\hline
Dataset & Usable videos & Total temporal images & Generated from deepfake videos \\\hline
UADFV           & 50   & 17,485    & 8,574     \\\hline
Celeb-DF        & 1164 & 2,44,022  & 1,52,262  \\\hline
FaceForensics   & 1990 & 6,36,391  & 3,18,217  \\\hline
DFD (Google)    & 2629 & 15,47,558 & 13,59,024 \\\hline
Celeb-DF-v2     & 6414 & 11,74,577 & 9,96,002  \\\hline
FaceForensics++ & 5925 & 16,36,474 & 13,29,329 \\\hline
\end{tabular}
\end{center}
\end{table}

\begin{figure}
    \centering
    \includegraphics[width=\textwidth]{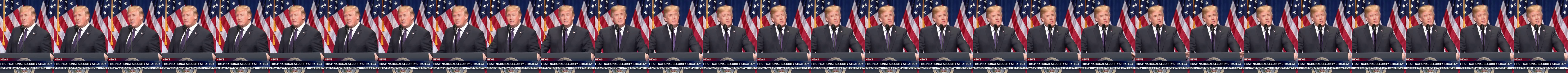}
    \caption{First 29 frames skipping every \nth{7} frame}
    \label{First 29}
\end{figure}

Figure \ref{First 29} shows the first 29 frames taken from video "0045\_fake" of UADFV dataset for the generation of a temporal image, skipping every \nth{7} frame.

\begin{figure}
    \centering
    \includegraphics[width=\textwidth]{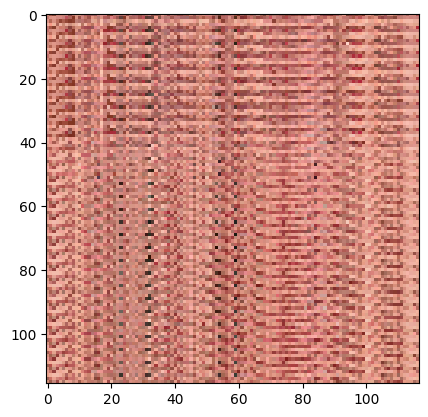}
    \caption{A temporal image of dimensions (116, 117, 3)}
    \label{116x117}
\end{figure}

Figure \ref{116x117} shows the temporal image generated using the 29 frames shown in \ref{First 29}.

As a large fraction of videos from the UADFV dataset had been skipped, it was no longer meaningful to use the temporal images generated from that dataset for training classification models.

Therefore, we further continued to generate temporal images of dimensions (78, 78, 3) to consider a much smaller duration of 2.833 seconds or 85 frames ($1 + (13-1)*7$) in order to include more videos. These temporal images used 6 rows for representing the 468 facial landmarks from each frame and hence includes only 13 frames ($13*6=78$) skipping every \nth{7} frame. Table \ref{Datasets 78x78} provides the details of the temporal images of dimensions (78, 78, 3) generated from each of the 6 different datasets.

\begin{table}[ht]
\begin{center}
\caption{Details of generated temporal for dimensions (78, 78, 3)}
\label{Datasets 78x78}
\begin{tabular}{|c|c|c|c|}
\hline
Dataset used    & Usable videos & Total temporal images & Generated from deepfake videos \\\hline
UADFV           & 98            & 25,947                & 12,806                         \\\hline
Celeb-DF        & 1,182         & 3,76,174              & 2,38,925                       \\\hline
FaceForensics   & 1994          & 8,59,707              & 4,29,828                       \\\hline
DFD (Google)    & 2715          & 18,51,102             & 16,29,301                      \\\hline
Celeb-DF-v2     & 6,497         & 19,01,639             & 16,23,809                      \\\hline
FaceForensics++ & 5933          & 23,01,275             & 18,83,204                      \\\hline
\end{tabular}
\end{center}
\end{table}

\begin{figure}
    \centering
    \includegraphics[width=\textwidth]{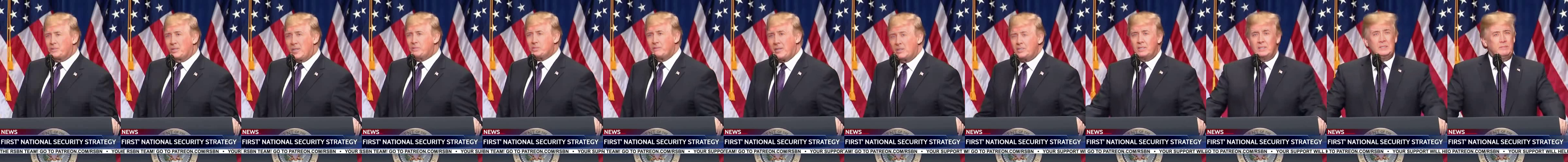}
    \caption{First 13 frames, skipping every \nth{7} frame}
    \label{First 13}
\end{figure}

Figure \ref{First 13} shows the first 13 frames taken from video “0045\_fake” of UADFV dataset for the generation of a temporal image of dimensions (78, 78, 3), skipping every \nth{7} frame.

\begin{figure}
    \centering
    \includegraphics[width=\textwidth]{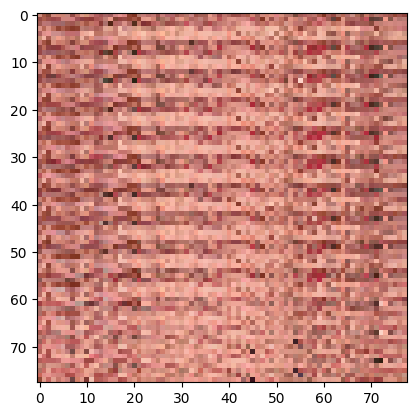}
    \caption{A temporal image of dimensions (78, 78, 3)}
    \label{78x78}
\end{figure}

Figure \ref{78x78} shows the temporal image generated using the 13 frames shown in \ref{First 13}.

\begin{figure}
    \centering
    \includegraphics[width=\textwidth]{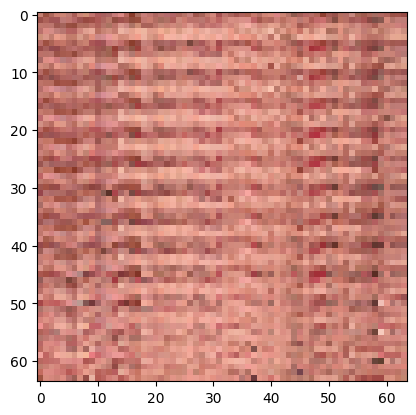}
    \caption{A temporal image of dimensions (78, 78, 3) resized to (64, 64, 3)}
    \label{78x78 to 64x64}
\end{figure}

Figure \ref{78x78 to 64x64} shows the temporal image from figure \ref{78x78} when resized to (64, 64, 3).

\begin{figure}
    \centering
    \includegraphics[width=\textwidth]{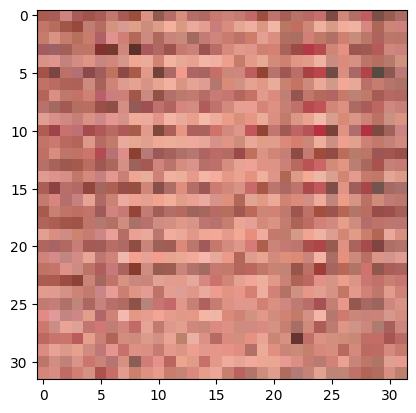}
    \caption{A temporal image of dimensions (78, 78, 3) resized to (32, 32, 3)}
    \label{78x78 to 32x32}
\end{figure}

Figure \ref{78x78 to 32x32} shows the same temporal image from figure \ref{78x78} when resized to (32, 32, 3).

\subsection{Formulations for selecting values of H,W and I}
Let $H$ be the height and $W$ be the width of the temporal image respectively. The values for $H$ and $W$ must be selected such that they satisfy the following two constraints:

\begin{enumerate}
\item
$\exists k : (kW = 468) \& (k \in \mathbb{N})$
\item
$H - W$ is small
\end{enumerate}

The first constraint ensures the width, $W$ is a factor of 468 so that all the pixels in each row of the temporal image would come from the same frame, whereas the second constraint ensures that the temporal image has a square shape, as the ImageNet models have been pre-trained with input images having equal height and width.

Let $S$ be the window size $: S = H/k$

Then the relationship between $H$, $W$ and $S$ can be formulated as: $$S = HW/468$$

We can see that the value of $S$ is dependent on the values of $H$ and $W$.
\\\\
For generalisation we can replace the value of 468 in the above constraints and formulas with a variable, say D, representing the constant number of interested data points  in each frame
\\\\
Let $I$ be the number of frames to skip, then the value of $I$ is given by the below formula:
\begin{equation}
\label{I equation}
   I = \lceil(X - 1)/(S - 1)\rceil
\end{equation}

The above equation comes from the relation: 
$$1 + (S - 1)I = X : I \in \mathbb{N}$$

Where $X$ represents the total number of frames covered by the sliding window

The values chosen for $H$ and $W$ were 78 and 78. The value for $S$ was thus 13. In order to select the value of $I$, we considered the average duration of a single eye blink and the mean blink rate for humans. 

The mean blink rate at rest is 17 blinks/min as per a study conducted in 1997. The blink rate was found to increase to 26 during conversation and drop as low as 4.5 while reading. \cite{bentivoglio1997analysis}

\begin{table}[ht]
\begin{center}
\caption{Calculating value of $I$ as per blink rate}
\label{blink rate table}
\begin{tabular}{|l|c|c|c|}
\hline
Blink rate (blinks/min)                   & 4.5    & 17               & 26             \\\hline
Blink rate (blinks/second)                & $4.5/60$ & $17/60$            & $26/60$          \\\hline
Occurrence of 1 blink in (seconds)        & $60/4.5$ & $60/17$            & $60/26$          \\\hline
Occurrence of 1 blink in X frames (30fps) & 400    & 105.88 $\sim$106 & 69.23 $\sim$70 \\\hline
Value of $I$ to reach X frames (S = 29)     & 15     & 4                & 3             \\\hline
\end{tabular}
\end{center}
\end{table}

As per the calculations shown in Table \ref{blink rate table}, the value of $I$ must be between 3 and 15 in order to include a frame in which the person of interest is blinking.
\\\\
The average duration of a single-eye blink is 0.1-0.4 sec. \cite{taschenbuch2001sensation} Therefore a single-eye blink will span across 3 to 12 frames for a 30 fps video and the maximum value $I$ can take is 12 without skipping over eye blinks.

Therefore We can rewrite the above two constraints for the value of $I$ as:
\begin{enumerate}
    \item 
    $3 <= I <= 15$ and
    \item
    $0 <= I <= 12$
\end{enumerate}

Therefore, 

\begin{equation}
\label{I constraint}
3 <= I <= 12
\end{equation}

Therefore we can conclude that the valid range for the value of $I$ will be between 3 and 12. Our previously chosen value of 7 is valid across this range, hence 7 was kept as the value of $I$.
\\\\
Combining equations \ref{I equation} and \ref{I constraint}, the final formula for choosing the value of $I$ becomes:
$$I = \lceil(X - 1)/(S - 1)\rceil : 3 <= I <= 12$$
\section{Findings and analysis}
\begin{figure}
    \centering
    \includegraphics[width=\textwidth]{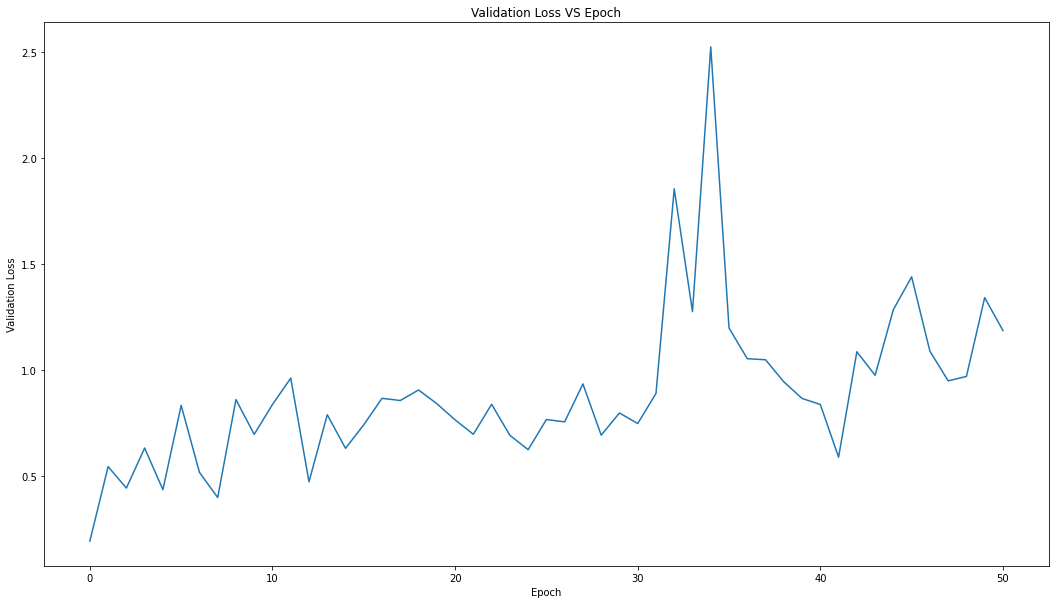}
    \caption{Learning curve for validation loss obtained after initial experiment}
    \label{Initial experiment}
\end{figure}

In our initial experiment we trained a pre-trained MobileNet model using the train and validation sets of the pre-processed temporal image dataset of dimensions (78, 78, 3) that had  been generated from the UADFV dataset with batch size set to 64 and learning rate set to $10^{-4}$. The test accuracy was 0.76201. The learning curves for validation accuracy and validation loss were noisy and a major point to note here was that the validation loss was lowest in the first epoch and was found to increase with each epoch, as can be seen in figure \ref{Initial experiment}. It was interpreted as the model diverging from the solution and hence a smaller batch size of 32 along with a different learning rate were decided to be used further on.

In an attempt to find the optimal learning rate and develop a better understanding, the experiment was re-run for different learning rates keeping the model as MobileNet. MobileNet had been chosen for its smaller size as it could yield quick results. Table \ref{Learning rates table} summarises our findings.

\begin{table}[ht]
\begin{center}
\caption{Results with different learning rates for MobileNet, batch size = 32}
\label{Learning rates table}
\begin{tabular}{|c|c|c|c|}
\hline
Learning Rate & Total Epochs & Test Accuracy & Test Loss \\\hline
1E-1          & 53           & 0.32657       & 0.77989   \\\hline
1E-2          & 82           & 0.94965       & 1.20005   \\\hline
1E-3          & 98           & 0.97319       & 0.58913   \\\hline
1E-4          & 128          & 0.73651       & 3.41414   \\\hline
1E-5          & 53           & 0.73847       & 3.66086   \\\hline
\end{tabular}
\end{center}
\end{table}

The model was underfitting for a learning rate of 0.1 whereas it showed good performance with learning rates of $10^{-2}$ and $10^{-3}$. $10^{-2}$ was chosen as the value for learning rate to be used for further experiments, as a larger learning rate is preferable to yield quick results and also since a learning rate of $10^{-3}$ may not work with all models.

 So far we have worked with only un-resized temporal images, therefore in order for testing the validity of resizing temporal images we conducted two experiments one with temporal images resized to (64, 64) and another where the temporal images were resized to (32, 32), keeping batch size as 32 and learning rate as $10^{-2}$. Table \ref{resized table} summarises our findings
 
\begin{table}[ht]
\begin{center}
\caption{Test results with resized temporal images}
\label{resized table}
\begin{tabular}{|c|c|c|c|}
\hline
Resized dimensions & Total Epochs & Test Accuracy & Test Loss \\\hline
(64, 64, 3) & 59 & 0.90323 & 0.93569 \\\hline
(32, 32, 3) & 53 & 0.72147 & 6.56100 \\\hline
\end{tabular}
\end{center}
\end{table}

\begin{figure}
    \centering
    \includegraphics[width=\textwidth]{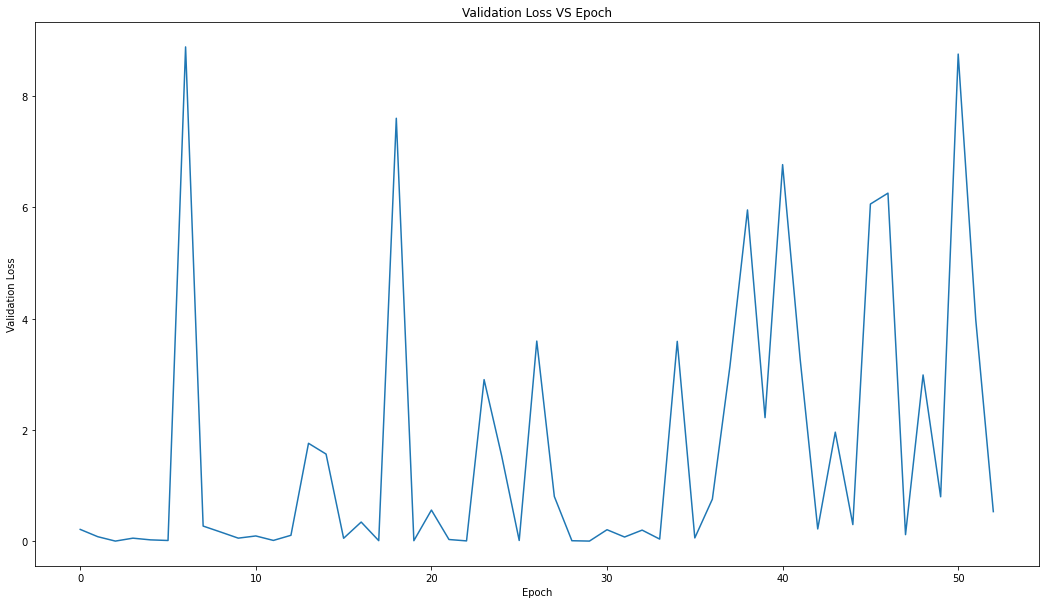}
    \caption{Learning curve for validation loss obtained on training MobilNet with temporal images resized to (32, 32, 3)}
    \label{val loss 32x32}
\end{figure}

In case of the temporal images resized to (32, 32, 3), the learning curve obtained for validation loss was extremely noisy as shown in figure \ref{val loss 32x32}.

The test accuracy was 0.71820 when the experiment was re-run with a batch size of 16, and the learning curve for validation accuracy was still noisy. Keeping the batch size as 32, when tried with a learning rate of 1E-1 the test accuracy was only 0.56755. Therefore we can infer that resizing of temporal images does not work unless it is close to the original dimensions.

In order to check whether transfer learning was helping, we trained a MobileNet model initialised with random weights using the un-resized temporal image dataset, keeping the batch size as 32 and learning rate as $10^{-2}$. The test accuracy was only 0.85060 compared to 0.94965 with transfer learning. Therefore it was inferred that transfer learning was indeed helping models to achieve a better performance.

We further decided to train different ImageNet models keeping the batch size as 32 and learning rate as $10^{-2}$. The results obtained have been provided in table \ref{table without restoring weights}.

InceptionV3 was the first model tried other than MobileNet. The learning curves for InceptionV3 were much smoother, although showing a very steep curve depicting an increase in accuracy or a decrease in loss during the initial epochs. We could possibly obtain better results from the model by tweaking the hyper-parameters, mainly patience (the number epochs to continue training the model in search of a new global optimum value. The training is stopped only if a new global optimum value was not found), or by restoring weights from the best epoch. Here the hyper-parameters have been chosen for optimal performance of the MobileNet model from our initial experiments. Patience had been set to 50 and we were not restoring weights from the best epoch.

Restoring weights from the best epoch was tried for Xception, and the test accuracy improved considerably to 97\%. from previously found 83\%. Therefore, it was inferred that weights from the best epoch should be restored for models with smoother learning curves, as a solution to overfitting. We further decided to train each of the models twice, once with restoring weights from best epoch and once without. Tables \ref{table without restoring weights} and \ref{table with restoring weights} summarise our findings and both tables have been sorted in descending order of test accuracy.

\begin{table}[ht]
\begin{center}
\caption{Test results without restoring weights from best epoch for all models}
\label{table without restoring weights}
\begin{tabular}{|c|c|c|c|}
\hline
Model             & Total Epochs & Test Accuracy & Test Loss \\\hline
ResNet50          & 54           & 0.96077       & 2.64214   \\\hline
MobileNetV2       & 87           & 0.95325       & 0.56588   \\\hline
MobileNet         & 82           & 0.94965       & 1.20005   \\\hline
DenseNet121       & 73           & 0.94704       & 2.34670   \\\hline
InceptionV3       & 94           & 0.90519       & 1.36978   \\\hline
InceptionResNetV2 & 61           & 0.90029       & 1.12898   \\\hline
EfficientNetB0    & 51           & 0.87577       & 3.36155   \\\hline
Xception          & 55           & 0.83360       & 3.41384   \\\hline
VGG16             & 102          & 0.32657       & 0.76874   \\\hline
VGG19             & 131          & 0.32657       & 0.73889   \\\hline
\end{tabular}
\end{center}
\end{table}

\begin{table}[ht]
\begin{center}
\caption{Test results with restoring weights from best epoch for all models}
\label{table with restoring weights}
\begin{tabular}{|c|c|c|c|}
\hline
Model             & Best Epoch & Test Accuracy & Test Loss \\\hline
MobileNetV2       & 27         & 0.97899       & 0.57454   \\\hline
Xception          & 10         & 0.97842       & 0.23335   \\\hline
InceptionResNetV2 & 51         & 0.94834       & 0.54508   \\\hline
InceptionV3       & 17         & 0.94508       & 0.39641   \\\hline
DenseNet121       & 5          & 0.85550       & 4.38512   \\\hline
EfficientNetB0    & 2          & 0.84766       & 0.93234   \\\hline
MobileNet         & 9          & 0.84700       & 2.15226   \\\hline
ResNet50          & 4          & 0.82183       & 1.50126   \\\hline
VGG16             & 50         & 0.32657       & 0.73964   \\\hline
VGG19             & 2          & 0.32657       & 0.72227   \\\hline
\end{tabular}
\end{center}
\end{table}

The test accuracies for the models; InceptionResNetV2, InceptionV3 and Xception were found to increase considerably by restoring weights from best epoch, with a very drastic improvement observed in case of Xception. For the models; DenseNet121, MobileNet, ResNet50 the test accuracies decreased considerably by restoring weights from best epoch. Models; EfficientNetB0 and MobileNetV2 showed similar performance under both conditions. Models; VGG16 and VGG19 performed terribly under both conditions. 

It is important to note that here the model was trained twice and the results shown in tables \ref{table without restoring weights} and \ref{table with restoring weights} have been obtained separately. By doing so we were able to confirm whether the observations made were a single time phenomenon or not.

VGG16 was underfitting, training accuracy was stuck at 0.5793 while validation accuracy was stuck at 0.2443 during  both attempts. The model also showed a very high training loss of the order of $10^{15}$ after the first epoch which further reduced to the order of $10^{-1}$ after further epochs. The same observations were made for VGG19 as well, although initial training loss was of the order of $10^{18}$ . Thus we can infer that the VGG models are incapable of recognizing the patterns needed for classification between temporal images generated from real and deepfake videos under these conditions. A curious thing to note here was that both VGG16 and VGG19 gave the same test accuracy of 0.32657 with slightly different test losses in all cases.

It is important to note that these models could give better performance under different hyper-parameters that are uniquely suitable to each of these models. For example, during our initial experiments we were able to find that the test accuracy shown by MobileNet increased to 97\% with a learning rate of $10^-3$. But when we further tried the learning rate of $10^-3$ with MobileNetV2 the test accuracy decreased considerably to 0.828375 from 0.95325 with $10^-2$ as learning rate. Therefore it was inferred that $10^-2$ was the optimal learning rate.

\begin{figure}
    \centering
    \includegraphics[width=\textwidth]{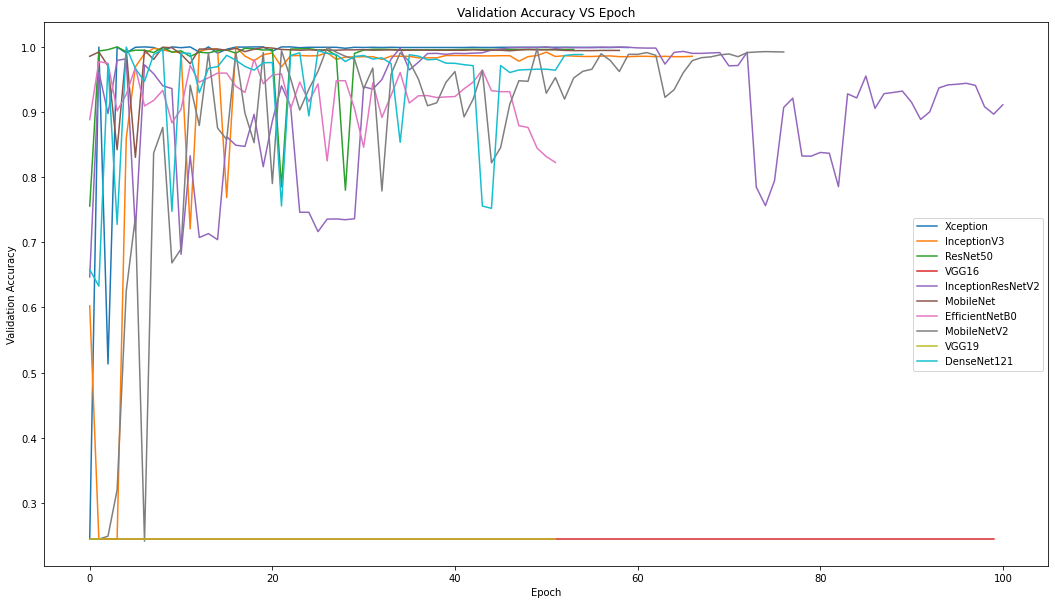}
    \caption{Learning curves for validation accuracy for all models on attempt with restoring weights from best epoch}
    \label{figure all models with}
\end{figure}

\begin{figure}
    \centering
    \includegraphics[width=\textwidth]{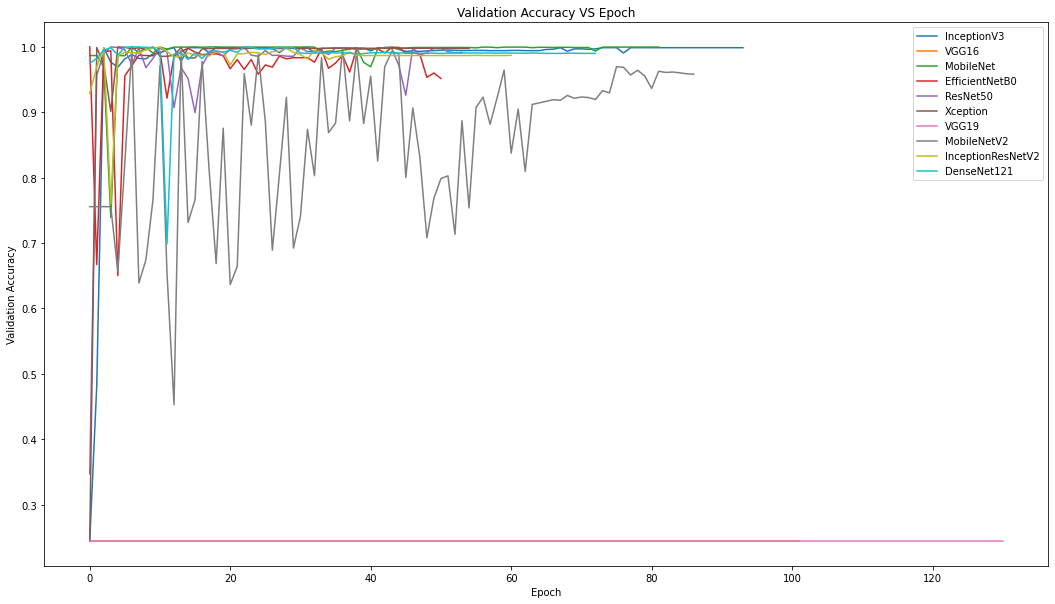}
    \caption{Learning curves for validation accuracy for all models on attempt without restoring weights from best epoch}
    \label{figure all models without}
\end{figure}

Figures \ref{figure all models with} and \ref{figure all models without} show the learning curves for validation accuracy for all models on attempts with and without restoring weights from best epoch respectively.

\section{Discussion}
 Thus we were able to successfully build and check the validity of temporal images using multiple ImageNet models and for several different hyperparameters. We were also able to formulate constraints that would aid further researchers to choose valid values for the height and width of temporal images and for the number of frames to skip. It was found that the dataset must be split into train, validation and test sets at the video level and not at the frame/temporal image level to avoid overfitting the model on the given dataset. MobileNet having a smaller number of parameters was chosen for our initial experiments as it could yield quick results. $10^{-2}$ was found to be the optimal learning rate along with a batch size of 32. We also found out that resizing of temporal images did not work unless the new dimensions were close to the original dimensions. Transfer learning was found to help models to achieve a better performance. We also inferred that weights from the best epoch should be restored for models with smoother learning curves, as a solution to overfitting. MobileNetV2 showed the overall best performance. Models; Xception, ResNet50, MobileNet, InceptionResNetV2, InceptionV3 and DenseNet121 also showed promising performances. Meanwhile, VGG16 and VGG19 were both found to underfit the temporal images generated from the UADFV dataset.
\section{Future scope}
Further advancements of the project include training and testing models on datasets other than UADFV and trying out variations of temporal images that are generated using only a fraction of the 468 facial landmarks, keeping particular interest to the facial landmarks from the eyes and mouth.
\section*{Acknowledgment}
The thesis was completed in partial fulfilment for the award of the degree of Bachelor of Technology in Computer Science and Engineering with Specialization in Bioinformatics at Vellore Institute of Technology (VIT). I would also like to express my gratitude to my guide, Dr. Sharmila Banu K, for allowing me to work on the topic of my choice.
\bibliographystyle{unsrt}
\bibliography{refs}
\end{document}